\documentclass[times,preprint,10pt]{elsarticle}

\usepackage{amsmath,amssymb}
\usepackage{graphicx}
\usepackage{booktabs,multirow,array}
\usepackage{makecell}
\usepackage{xurl}
\usepackage{placeins}

\biboptions{sort&compress}
\graphicspath{{figs/}{./}}
\DeclareGraphicsExtensions{.pdf,.png,.jpg,.jpeg}
\providecommand{\tightlist}{\setlength{\itemsep}{0pt}\setlength{\parskip}{0pt}}

\journal{Pattern Recognition}

\begin{document}

\begin{frontmatter}

\title{Noise-Robust Box-Supervised Infrared Small Target Detection via Physics-Inspired Soft Label Optimization}

\author[3]{Xizhe Zhang}
\author[1,2,3]{Fan Shi\corref{cor1}}
\ead{shifan@email.tjut.edu.cn}
\author[1,2]{Mianzhao Wang}
\author[3]{Jiangpeng Zheng}
\author[3]{Xu Cheng}
\author[1,2,3]{Shengyong Chen}

\affiliation[1]{organization={Engineering Research Center of Learning-Based Intelligent System, Ministry of Education, Tianjin University of Technology}, city={Tianjin}, postcode={300384}, country={China}}
\affiliation[2]{organization={Key Laboratory of Computer Vision and System, Ministry of Education, Tianjin University of Technology}, city={Tianjin}, postcode={300384}, country={China}}
\affiliation[3]{organization={School of Computer Science and Engineering, Tianjin University of Technology}, city={Tianjin}, postcode={300384}, country={China}}
\cortext[cor1]{Corresponding author}

\begin{abstract}
Infrared small target detection (IRSTD) commonly relies on pixel-level mask supervision. Such annotations, however, are costly and inherently uncertain because infrared targets have blurred boundaries and weak textures. We formulate box-supervised IRSTD as a problem distinct from generic box-to-mask segmentation and point-supervised IRSTD. Its central challenge is to construct stable pixel-level soft supervision from highly contaminated boxes. To this end, we propose Hotspot-Anchored Label Optimization (HALO). HALO localizes a radiometric anchor inside each box under local background-statistics constraints, then synthesizes a Physically Anchored Gaussian (PAG) soft label around the anchor. This turns noisy box supervision into continuous, pixel-level soft labels. The entire process is performed offline before training, remains decoupled from the detector backbone, and requires no online label updates. Experiments on public datasets show that HALO is competitive with representative box-supervised methods under standard tight boxes. Under looser or shifted box annotations that better approximate real scenarios, HALO is substantially more robust while remaining consistent across backbones. We further introduce a contamination-aware operating-regime analysis to characterize the effective boundary of this class of methods and reveal how intrinsic signal-to-clutter ratio relates to performance.
\end{abstract}

\begin{keyword}
Infrared small target detection \sep Box supervision \sep Soft label \sep Weakly supervised segmentation \sep Signal-to-clutter ratio
\end{keyword}

\end{frontmatter}

\section{Introduction}\label{sec:introduction}

Infrared small target detection (IRSTD) is a core task in infrared
search and tracking, with applications in traffic monitoring, maritime
rescue, and military reconnaissance \cite{cheng2024review}. Although
termed detection, the task is usually formulated as pixel-level
segmentation. It is challenging because infrared targets are extremely
small and weakly textured, often occupying only a few pixels with limited
contrast against cluttered backgrounds \cite{rawat2020review}. They are
also sparsely distributed, which creates a severe foreground-background
imbalance and leaves little local radiometric separability, so genuine
target responses are easily buried in clutter. Casting the task as
segmentation lets models exploit shape and contrast cues, but it also
demands dense pixel-level masks for training.

Existing IRSTD methods fall into fully supervised and weakly supervised
paradigms. Fully supervised methods such as DNANet \cite{li2023dnanet}
and UIU-Net \cite{wu2023uiunet} achieve strong performance but rely on
pixel-level masks, which are costly to annotate and often ambiguous for
small targets with blurred boundaries and weak appearance cues. This
motivates weaker but more accessible supervision, where annotations trade
labeling effort against label completeness and different forms sit at
different points of this trade-off. Weakly supervised IRSTD has so far
mainly followed the point-supervision paradigm, e.g., LESPS
\cite{ying2023lesps} and PAL \cite{yu2025pal}. Point annotations require
only a center or approximate location, which greatly reduces cost, but
they carry no extent information, so these methods must grow each point
into a region through pseudo-mask generation or network-in-the-loop label
evolution before training.

\begin{figure}
\centering
\includegraphics[width=\textwidth]{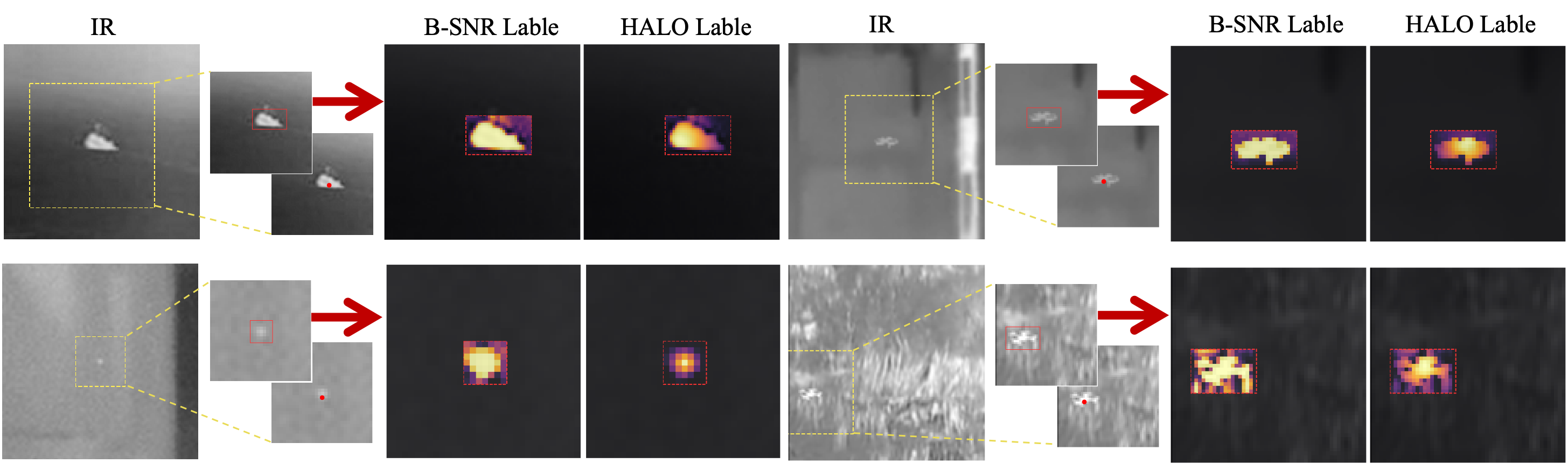}
\caption{Representative comparison between box and point supervision. Each case shows the infrared scene with a zoomed-in region, the intermediate radiometric response map, and the final HALO soft label.}
\label{fig:boxpoint}
\end{figure}

Box supervision is another common weak-supervision form, though not a
direct extension of point-supervised IRSTD. A point gives a sparse
location prior, whereas a box gives a coarse region prior that, for
infrared small targets, is usually dominated by background and may suffer
from localization errors \cite{xiao2025abrnet}. Pseudo-label generation
thus shifts from expanding sparse cues into masks toward purifying a
contaminated region. The goal becomes to localize the true target
response inside a noisy box and turn it into stable soft supervision.
This makes box-supervised IRSTD a dedicated problem setting.
Fig.~\ref{fig:boxpoint} illustrates the distinction and shows how HALO
converts a noisy box into pixel-level soft supervision.

Research on box-level supervision for IRSTD remains limited. Generic
box-to-mask methods for natural images, such as GrabCut
\cite{rother2004grabcut} and BoxInst \cite{tian2021boxinst}, rely on
texture, edge, or appearance-consistency assumptions. They broadly fall
into two families, one driven by foreground-background appearance
separability and one by geometric or energy constraints, and both
presuppose cues that infrared lacks. In IRSTD these assumptions break
down. A target is a weak local hotspot with blurred boundaries and low
contrast, lacking the shape and appearance such cues require.
Box-supervised IRSTD needs a supervision-generation framework tailored to
infrared imaging characteristics.

To the best of our knowledge, no dedicated framework has used infrared
imaging priors to generate box supervision for IRSTD. We propose
Hotspot-Anchored Label Optimization (HALO), a physics-inspired
box-to-soft-label framework. Its design exploits a simple physical fact:
an infrared target is a compact thermal hotspot whose energy spreads over
a few pixels through the sensor's point-spread function, leaving a stable
local intensity peak even where the boundary is ill-defined. Unlike
methods that fill boxes as foreground
or rely on online label evolution, HALO constructs soft supervision
offline from the locally identifiable radiometric structure of infrared
targets: it first localizes a radiometric anchor under local
background-statistics constraints, then builds a Physically Anchored
Gaussian (PAG) soft label around it, turning highly noisy boxes into
continuous soft supervision for training. Because the anchor is defined
by the strongest local radiometric response inside the box, it is largely
insensitive to the box boundary, so HALO stays robust under loose or
shifted boxes; we confirm this experimentally, especially at low
signal-to-clutter ratio (SCR). On NUAA-SIRST, NUDT-SIRST, and IRSTD-1k,
HALO is competitive with box-supervised baselines under tight boxes and
degrades far more gracefully under loose or shifted ones. Beyond accuracy,
we relate this behavior to a measurable image-quality statistic through a
contamination-aware operating-regime analysis, linking intrinsic SCR to
when such offline soft labels stay reliable, an angle not previously
examined for weakly supervised IRSTD. The main contributions of this
paper are as follows:

\begin{enumerate}
\def\labelenumi{\arabic{enumi}.}
\tightlist
\item
  We formulate box-supervised IRSTD as an independent problem, distinct
  from generic box-to-mask segmentation and point-supervised IRSTD; its
  core difficulty is not boundary recovery but reliably localizing the
  target radiometric structure inside heavily contaminated boxes.
\item
  We propose HALO, which applies infrared imaging priors at the
  supervision-generation stage rather than in network design: through
  radiometric anchor localization and Physically Anchored Gaussian (PAG)
  soft-label synthesis, it converts noisy coarse boxes into trainable
  soft labels offline and decoupled from the backbone.
\item
  We quantify, for the first time in IRSTD, the overlooked dimension of
  box-annotation noise, showing that HALO is markedly more robust to
  realistic noisy boxes; we further introduce a contamination-aware
  operating-regime analysis that links the effective boundary of such
  methods to intrinsic dataset SCR, characterizing both the strengths and
  limitations of HALO.
\end{enumerate}

\section{Related Work}\label{sec:related_work}

\subsection{Infrared Small Target Detection and Weak
Supervision}\label{infrared-small-target-detection-and-weakly-supervised-irstd}

Early conventional methods mainly relied on handcrafted priors and fall
into three families. Filtering-based methods suppress the background and
threshold the residual, evolving from Max-Mean/Max-Median filters
\cite{deshpande1999maxmean} and morphological processing
\cite{rivest1996detection} to the Top-Hat operator \cite{zeng2006tophat}.
Local-contrast methods instead measure target saliency against a local
neighborhood, as in the local contrast measure \cite{chen2013local}, its
tri-layer-window variant \cite{han2019trilayer}, and the weighted
strengthened local contrast measure \cite{han2020weighted}. Low-rank
methods separate a sparse target from a low-rank background, including
the infrared patch-image model \cite{gao2013ipi}, its reweighted
patch-tensor extension with nonlocal and local priors
\cite{dai2017reweighted}, and a non-convex tensor low-rank approximation
\cite{liu2021nonconvex}. These methods detect targets by enhancing
statistical differences between the target and the background. However,
they often lack stable generalization when targets are extremely small,
background clutter is strong, or scene statistics change substantially.

In recent years, deep-learning-based segmentation methods have
substantially improved IRSTD performance. MDvsFA \cite{wang2019mdvsfa}
was among the early works to formulate Infrared Small Target Detection
as a segmentation problem. ACM \cite{dai2021acm} and ALCNet
\cite{dai2021alcnet} emphasize local contrast and context-modulation
priors. DNANet \cite{li2023dnanet}, UIU-Net \cite{wu2023uiunet}, and
RKFormer \cite{zhang2022rkformer} further strengthen extremely small
target representations through dense skip connections, multi-scale
feature fusion, and dedicated attention structures. More recently,
transformer-based methods, such as SCTransNet \cite{yuan2024sctransnet},
model long-range dependencies through spatial-channel cross-attention.
Other studies improve fully supervised IRSTD representations through
graph context modeling \cite{gclnet2024}, frequency-spatial fusion
\cite{fsfnet2024}, and low-level detail preservation \cite{li2025ilnet}.
Together, these methods establish strong fully supervised baselines, but
nearly all rely on dense pixel-level mask supervision. For IRSTD,
infrared small targets usually occupy only a few pixels and have blurred
boundaries. Mask annotation is therefore costly and highly uncertain.
The main bottleneck of high-performance IRSTD is gradually shifting from
whether networks can model the task to whether supervision signals can
be obtained at low cost.

Against this background, representative work on weakly supervised IRSTD
has recently focused on single-point supervision. These methods usually
start from the target center point or its approximate location. They
then obtain pixel-level supervision for training through pseudo-mask
generation or label evolution. According to the pseudo-label generation
strategy, they can be broadly divided into two categories. One category
constructs local regions around points and recovers pseudo masks using
rule-based strategies, energy functionals, clustering, or size priors,
as in COM \cite{li2024levelset} and MCLC \cite{li2023mclc}. The other
category relies on network-in-the-loop iterative label updates. It
gradually expands point supervision through network predictions, as in
LELCM \cite{yang2024lelcm}, LESPS \cite{ying2023lesps}, and PAL
\cite{yu2025pal}. HMG \cite{he2025hmg} lies between these categories by
combining rule-based generation with network refinement. Given the
representative status of LESPS \cite{ying2023lesps} and PAL
\cite{yu2025pal}, we use them as key reference methods in the
experimental section.

Existing point-supervised methods provide important references for
weakly supervised IRSTD. However, point supervision corresponds to
sparse location priors, whereas box supervision provides region priors
with substantial background contamination. The two settings therefore
face different pseudo-label generation difficulties. When
bounding-box annotations are already available, transforming them
into stable soft supervision for segmentation training becomes a
distinct problem from existing point-supervision routes.

\subsection{Box Supervision and Box-to-Soft-Label
Generation}\label{box-supervision-and-box-to-soft-label-generation}

In general vision tasks, box supervision is widely used for weakly
supervised segmentation. Existing methods can be broadly divided into
two categories. The first category emphasizes appearance separability or
region refinement during training. GrabCut \cite{rother2004grabcut}, for
example, recovers object masks inside boxes using graph cuts and
foreground-background appearance separability. BoxSup
\cite{dai2015boxsup} uses unsupervised region proposals as candidate
masks and iteratively refines them against the box constraint during
training. BBAM \cite{lee2021bbam}
generates finer-grained object regions using detector attribution maps.
Although these methods differ in implementation, they all rely on
statistically distinguishable foreground-background differences in
texture, color, or appearance response. The second category is based on
geometric constraints and energy minimization. BoxInst
\cite{tian2021boxinst} approximately recovers target regions using
box-induced horizontal/vertical projection consistency, complemented by
a low-level color-similarity term between neighboring pixels. Box2Mask \cite{li2022box2mask} embeds level-set
evolution into deep networks and iteratively optimizes instance masks
inside boxes by minimizing the Chan-Vese energy function. DiscoBox
\cite{lan2021discobox}, Tightness Prior \cite{hsu2019tightness},
BoxSnake \cite{yang2023boxsnake}, and BoxSeg \cite{lai2025boxseg}
typically rely on relatively clear boundaries or appearance/color cues. Overall, these
methods rely on rich texture, clear boundaries, or strong
foreground-background appearance separability. These conditions are
often difficult to satisfy in IRSTD because infrared small targets
occupy very few pixels, have weak textures and blurred boundaries, and
exhibit low background contrast. Such methods are therefore sensitive to
box quality. When boxes are loose or shifted, substantial in-box
background can easily be included as foreground. Generic box-to-mask
methods thus have limited robustness under realistic noisy box
annotations. This motivates supervision-generation mechanisms tailored
to infrared imaging characteristics and robust to box errors.

\subsection{Imaging Physics and Priors in Infrared
Perception}\label{physical-priors-in-infrared-perception}

Unlike ordinary natural-image segmentation, infrared small target
detection is constrained by imaging physics and local background
statistics. The spatial distribution of target energy on the image plane
often follows a diffusion pattern. Local statistical methods represented
by constant false alarm rate (CFAR) \cite{rohling1983cfar} have long
been used for infrared target detection. Their core idea is to measure
target saliency through local background mean and variance estimation.
Point spread function (PSF) \cite{moradi2016psf}, full width at half
maximum (FWHM) \cite{rogalski2016challenges}, and sub-pixel energy spread
\cite{ma2021infrared} describe the spatial diffusion behavior of small
targets in imaging systems. From a broader research perspective,
traditional local contrast methods
\cite{chen2013local} and deep models
such as ACM \cite{dai2021acm} and ALCNet \cite{dai2021alcnet} share a
similar infrared perception intuition. Target saliency depends on local
statistical differences. However, these priors have mostly been used for
detection-response modeling rather than explicit supervision generation.
Although infrared imaging priors are widely used in IRSTD network-architecture
design, recent work has also introduced local contrast statistics into
point-supervised pseudo-label generation \cite{yang2024lelcm}. To the
best of our knowledge, no prior work has specifically studied how to use
infrared imaging priors to establish stable training supervision from box
annotations. This is precisely the problem addressed by HALO.

In summary, existing work provides important background from fully
supervised modeling, point-supervised label evolution, generic
box-supervised segmentation, and infrared imaging priors. However,
these lines of work have not formed a unified solution path for
box-supervised IRSTD. This indicates that box-to-soft-label generation
remains a distinct problem in IRSTD and merits dedicated study.

\section{The Proposed HALO Framework}\label{sec:method}

\begin{figure}
\centering
\includegraphics[width=0.9\textwidth]{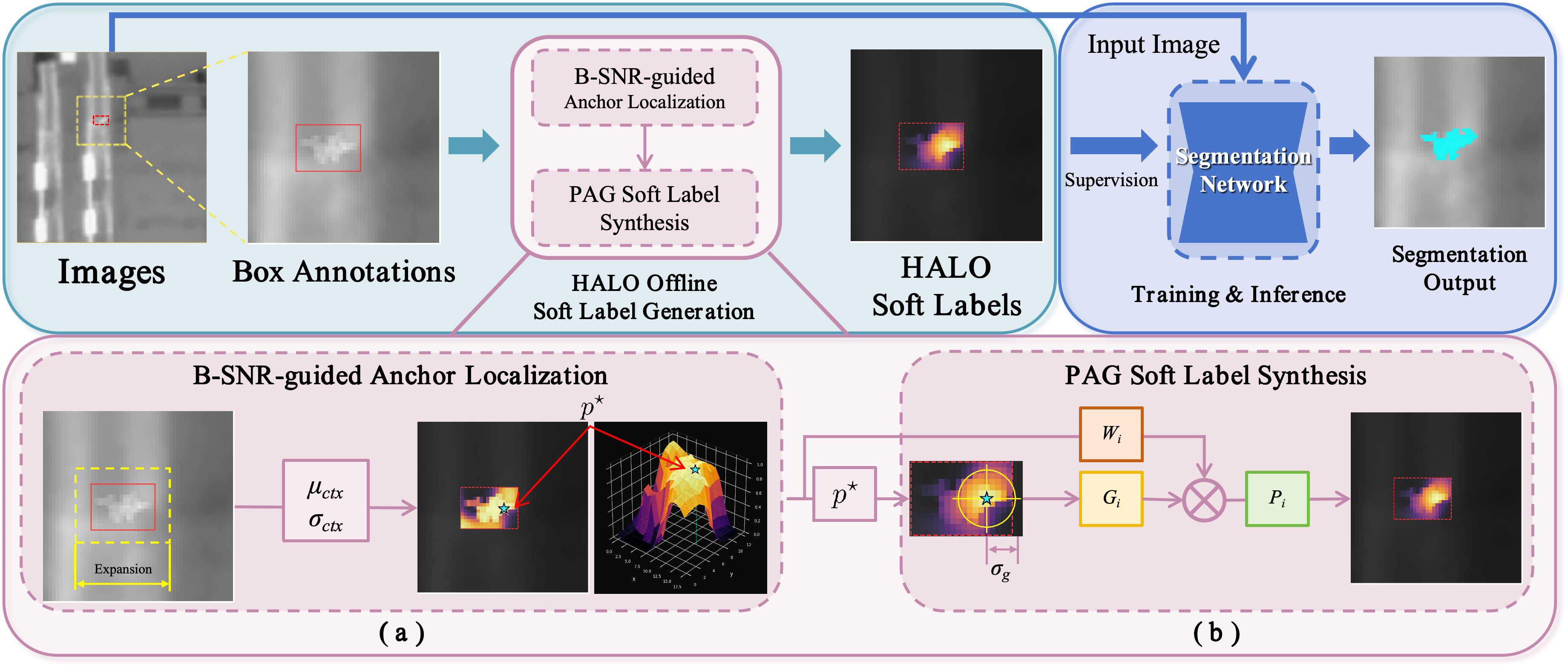}
\caption{Overall HALO framework: from infrared images and coarse boxes, through offline pseudo-label generation, to downstream detector training under soft label supervision and segmentation output. (a) Stage I performs background-normalized signal-to-noise ratio (B-SNR)-guided radiometric anchor localization and confidence field generation. (b) Stage II performs PAG soft-label generation.}
\label{fig:pipeline}
\end{figure}

We propose HALO for the box-supervised IRSTD problem studied in this
paper, where the goal is to generate trainable soft supervision from
coarse bounding boxes. The method consists of two stages. Fig.~\ref{fig:pipeline}(a)
shows Stage I, which localizes a reliable radiometric anchor inside a
coarse box and generates the corresponding confidence field. Fig.~\ref{fig:pipeline}(b)
shows Stage II, which constructs a continuous and smooth soft label
around the anchor. The main intermediate quantities and symbols are
introduced in the following subsections and kept consistent with the
figure annotations.

\subsection{Stage I: Background-Normalized Signal-to-Noise Ratio (B-SNR)-guided Radiometric Anchor
Localization}\label{stage-i-b-snr-guided-radiometric-anchor-localization}

In box-supervised IRSTD, the core issue introduced by a coarse box is
not boundary inaccuracy but uncertainty about the target location inside
the box. Stage I therefore localizes a physically meaningful radiometric
anchor inside the coarse box and generates the confidence field used for
subsequent soft-label construction.

HALO does not treat the entire box as foreground, nor does it use the
geometric center. Instead, it constructs a local context around the box
and uses the background statistics of this region to measure the
radiometric saliency of in-box pixels relative to the background.

Given a bounding box \(B=[x_1,x_2]\times[y_1,y_2]\), its width and
height are \(w=x_2-x_1\) and \(h=y_2-y_1\), respectively. To obtain a
stable local background reference, we construct a context region using a
scale-adaptive expansion strategy with a minimum constraint. The
expansion size is defined as

\[
p_x=\max\!\left(\left\lfloor \frac{r-1}{2}w\right\rfloor,3\right),\qquad
p_y=\max\!\left(\left\lfloor \frac{r-1}{2}h\right\rfloor,3\right),
\]

where \(r\) is the expansion ratio, set to \(r=1.5\) by default. The
variables \(p_x\) and \(p_y\) denote the numbers of expanded pixels in
the horizontal and vertical directions, respectively. This design
preserves scale adaptivity while imposing a lower bound of 3 pixels,
which avoids an overly small background window for extremely small
targets. The expanded context region is
\(\Omega_{ctx}=[x_1-p_x,\,x_2+p_x]\times[y_1-p_y,\,y_2+p_y]\). In
implementation, any part outside the image boundary is clipped. This
step is used only to stabilize background statistics and is not an
independent module.

We then compute local background statistics on \(\Omega_{ctx}\):

\[
\mu_{ctx} = \frac{1}{|\Omega_{ctx}|}\sum_{j\in\Omega_{ctx}} I_j,\qquad
\sigma_{ctx} = \sqrt{\frac{1}{|\Omega_{ctx}|}\sum_{j\in\Omega_{ctx}}(I_j-\mu_{ctx})^2},
\]

where \(I_j\) denotes the grayscale value of pixel \(j\). The terms
\(\mu_{ctx}\) and \(\sigma_{ctx}\) denote the local mean and standard
deviation, respectively. These statistics are estimated over the entire
expanded window rather than only the outside-box region, which provides
a more stable local contrast reference. Based on these statistics, the
B-SNR response of each in-box pixel \(i\in B\) is defined as

\[
s_i = \frac{I_i-\mu_{ctx}}{\sigma_{ctx}+\varepsilon},
\]

where \(\varepsilon\) is a numerical stabilization term, set to
\(10^{-4}\). To obtain stable soft weights, we apply a sigmoid
transformation with temperature parameter \(\tau\) to \(s_i\), yielding
\(\tilde W_i=\sigma(\tau s_i)\). The parameter \(\tau\) controls how
sharply the soft weight changes with B-SNR. A larger \(\tau\) makes the
transition between high-response suspected target pixels and
low-response background pixels sharper, whereas a smaller \(\tau\) makes
it smoother. We set \(\tau=3.0\) in this paper. We then perform min-max
normalization inside the box to obtain the final weight \(W_i\):

\[
W_i = \frac{\tilde W_i-\min_{j\in B}\tilde W_j}{\max_{j\in B}\tilde W_j-\min_{j\in B}\tilde W_j}.
\]

When the in-box responses are nearly identical, we directly set
\(W_i=0.5\). Finally, \(W_i\) forms the B-SNR confidence field,
characterizing the relative radiometric saliency of in-box pixels.

The radiometric anchor is defined as
\(p^\star = \arg\max_{i\in B} s_i\). Thus, Stage I outputs the
radiometric anchor \(p^\star\), the B-SNR response field \(s_i\), and the
confidence field \(W_i\) as inputs to Stage II.

\subsection{Stage II: Physically Anchored Gaussian Soft Label
Synthesis}\label{stage-ii-physically-anchored-gaussian-pag-soft-label-synthesis}

The anchor and confidence field obtained in Stage I are still
insufficient to form stable supervision directly. To avoid overly sparse
single-point supervision and instability from hard thresholding, Stage
II estimates the diffusion scale of the principal-energy region around
the anchor. It then constructs a spatial Gaussian distribution and fuses
it with the confidence field to generate a continuous soft label.

Let the anchor response be \(s_{p^\star}=s(p^\star)\). Following the
idea of full width at half maximum (FWHM), we define the half-maximum
principal-energy region as
\(\Omega_{1/2}=\{\,i\in B \mid s_i \ge 0.5\,s_{p^\star}\,\}\). The
circle in Fig.~\ref{fig:pipeline}(b) only illustrates this local support scale and does
not correspond to the final label boundary. We further define the
non-negative response \(s_i^+=\max(s_i,0)\) and estimate the adaptive
Gaussian scale as

\[
\sigma_g = \max\!\left(
\gamma \sqrt{
\frac{\sum_{i\in\Omega_{1/2}} s_i^+ \lVert i-p^\star\rVert^2}
{\sum_{i\in\Omega_{1/2}} s_i^+}
},
1.0
\right),
\]

where \(\gamma=1.5\). When the effective response inside the
principal-energy region is too weak, the scale falls back to
\(\sigma_g=1.0\). Here, \(\sigma_g\) is neither a pre-set empirical
width nor an exact recovery of the true PSF. Instead, it is adaptively
estimated from the spatial distribution of the principal-energy region
around the anchor. We then construct an anchor-centered spatial Gaussian
term
\(G_i=\exp\!\left(-\frac{\lVert i-p^\star\rVert^2}{2\sigma_g^2}\right)\)
and fuse it with the confidence field to obtain the final single-box
soft label:

\[
P_i = W_i G_i.
\]

This design satisfies both the radiometric-response constraint and the
spatial-structure constraint. For multiple boxes, we adopt pixel-wise
maximum fusion:

\[
Y_i^{\text{HALO}} = \max_k P_i^{(k)}.
\]

Because targets in IRSTD scenes are typically sparse, adjacent or
overlapping boxes are uncommon. For overlapping regions that do occur,
pixel-wise max preserves the stronger soft-label value from the
corresponding boxes. This avoids dilution caused by averaging and
prevents one target from suppressing another. When the Gaussian supports
of two anchors meet at their edges, max fusion also keeps their peaks
from weakening each other. This fusion rule is consistent with
single-box generation and introduces no additional parameters or
thresholds. Stage II thus converts the output of Stage I into the final
HALO soft label, which can be directly used for network training.

\subsection{Training Interface}\label{training-interface}

We adopt DNANet \cite{li2023dnanet} as the default downstream
segmentation backbone. This model performs strongly on IRSTD and is well
suited to modeling extremely small targets. It is therefore appropriate
for evaluating the quality of the supervision signals generated by HALO.

During training, the offline-generated HALO soft label replaces the
original pixel-wise annotation, while the network architecture and
optimization pipeline remain unchanged. During testing, the model
directly predicts the input infrared image without relying on bounding
boxes or the label-generation process. To verify the generality of the
method, we also conduct supplementary experiments on other segmentation
networks. The results are reported in the experimental section.

For the training objective, we use soft intersection-over-union (SoftIoU) loss to directly optimize the
overlap between the network prediction and the HALO soft label:

\[
\mathcal{L}_{\mathrm{SoftIoU}} =
1 - \frac{\sum_i \hat Y_i Y_i^{\text{HALO}} + \xi}
{\sum_i \hat Y_i + \sum_i Y_i^{\text{HALO}} - \sum_i \hat Y_i Y_i^{\text{HALO}} + \xi},
\]

where \(\hat Y_i=\sigma(z_i)\) is the prediction probability and
\(\xi=0.8\) is the smoothing term.

As an offline supervision scheme, HALO soft labels are generated once
before training and remain fixed throughout training. They therefore
cannot be gradually corrected by network feedback as in online methods.
This is a deliberate trade-off. The cost is that label quality depends
entirely on the generation stage and cannot self-correct during
training. The benefit is that HALO avoids the dependence of
network-in-the-loop methods, such as LESPS \cite{ying2023lesps} and PAL
\cite{yu2025pal}, on the current network state. It has no cold-start
problem, requires no repeated label updates or additional forward-pass
overhead, and is fully reproducible for the same input. Because HALO
soft labels are driven by infrared imaging priors rather than network
predictions, their quality does not drift with unstable predictions in
early training.

\begin{figure}
\centering
\includegraphics[width=0.95\textwidth]{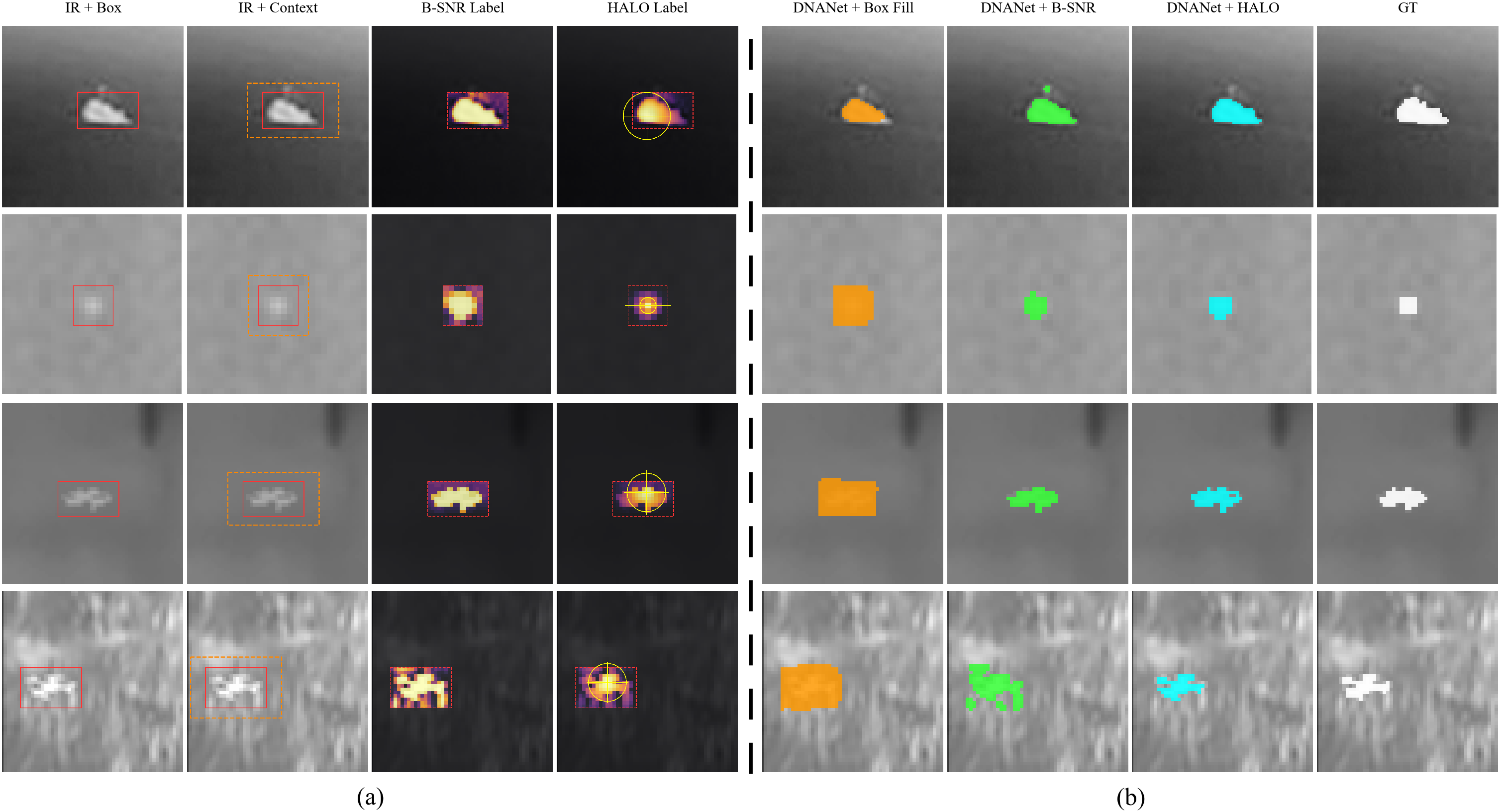}
\caption{Representative qualitative comparison results. (a) Intermediate process of generating supervision from coarse bounding boxes. (b) DNANet predictions trained with Box Fill, B-SNR, and HALO supervision, respectively, compared with ground truth.}
\label{fig:qualitative}
\end{figure}

Fig.~\ref{fig:qualitative} presents representative qualitative results. Fig.~\ref{fig:qualitative}(a) shows the
generation process from coarse boxes to B-SNR intermediate labels and
then to HALO soft labels. Fig.~\ref{fig:qualitative}(b) compares DNANet predictions trained
under Box Fill, B-SNR, and HALO supervision. HALO generates more
concentrated and smoother supervision, leading to predictions closer to
the true target regions.

\section{Experiments}\label{sec:experiments}

\subsection{Implementation Details}\label{implementation-details}

We evaluate HALO on three public infrared small target datasets:
NUAA-SIRST \cite{dai2021acm}, NUDT-SIRST \cite{li2023dnanet}, and
IRSTD-1k \cite{zhang2022isnet}. They contain 427, 1327, and 1001 images,
respectively. NUAA-SIRST and IRSTD-1k are real infrared datasets,
whereas NUDT-SIRST is synthesized from real infrared images.

The main experiments adopt a controlled box-supervision setting. Box
annotations are automatically converted from the original pixel-level
ground-truth (GT) masks by extracting the tight bounding rectangle of each
foreground connected component. Pixel-level annotations are used only to
construct the fully supervised upper bound and for final evaluation.
This tight-box protocol is designed to isolate the box-to-label
mechanism, rather than to model all possible annotation errors; the
effect of looser or shifted boxes is evaluated separately in the
robustness experiments. The protocol is consistent with common
evaluation practice in point-supervised IRSTD, such as LESPS
\cite{ying2023lesps} and PAL \cite{yu2025pal}. Weak annotations are not
manually re-collected, but automatically derived from complete
pixel-level annotations, which minimizes the effect of upstream
annotation-collection errors. Point annotations and box annotations do
not contain the same information: the former mainly provide a location
prior, whereas the latter also provide a coarse scale range. We therefore
include point-supervised methods in Table~\ref{tab:main_results} mainly
as a weak-supervision reference band. The comparison positions box
supervision within the weak-supervision spectrum, rather than claiming
strict competition under equal information content.

Unless otherwise stated, we use DNANet as the default backbone with a
ResNet-18 encoder. The network is trained for 1500 epochs using the
Adagrad optimizer and CosineAnnealingLR scheduler, with a batch size of
4 and an initial learning rate of 0.05. Training uses random scale
resizing, random cropping, random horizontal flipping, and mild Gaussian
blur. During testing, all images are resized to \(256\times256\). All
experiments use public dataset splits and are conducted with Python
3.9.20 and PyTorch 2.1.2.

HALO pseudo labels are generated offline before training. The
hyperparameters involved in HALO (\(r\), \(\tau\), \(\gamma\),
\(\varepsilon\), and \(\xi\)) are defined in the method section. They
are fixed constants chosen according to infrared imaging physics and
numerical stability, rather than learnable parameters tuned on the
validation set. The choices of \(r\) and \(\gamma\) are further examined
in the parameter-sensitivity analysis below. We use mean
intersection-over-union (mIoU) as the primary evaluation metric, with
probability of detection (\(P_d\)) and false-alarm rate (\(F_a\)) as
auxiliary detection metrics. For mIoU, the predicted probability map is
binarized at a fixed threshold of \(0.3\), applied identically to all
compared methods, and the mIoU with the ground-truth
mask is computed over the test set. \(P_d\) and \(F_a\) are evaluated at
a fixed probability threshold of \(0.5\), also applied consistently to
all methods. A target is counted as detected if the centroid
distance between a predicted connected component and the ground-truth
target is less than \(3\) pixels. \(P_d\) is the ratio of detected
targets to all ground-truth targets, and \(F_a\) is the ratio of
unmatched predicted pixels to all image pixels.

\subsection{Main Results and Mechanism
Validation}\label{main-results-and-mechanism-validation}

\paragraph{Main Results.}

Table~\ref{tab:main_results} reports results under different supervision
forms. The box-supervised methods and HALO are the main comparison;
point- and unsupervised methods form a weak-supervision reference band,
and the fully supervised result is the upper bound.

\begin{table}[t]
\caption{Main results under different supervision forms. Dataset columns report mean intersection-over-union (mIoU, \%) / probability of detection ($P_d$, \%) / false-alarm rate ($F_a$).}
\label{tab:main_results}
\centering
\footnotesize
\setlength{\tabcolsep}{2pt}
\renewcommand{\arraystretch}{1.2}
\resizebox{\textwidth}{!}{%
\begin{tabular}{@{}llcccc@{}}
\toprule
Supervision & Method & NUAA-SIRST & NUDT-SIRST & IRSTD-1k & Avg. mIoU (\%) \\
\midrule
\multirow{3}{*}{Unsupervised}
& TLLCM & 11.03 / 79.47 / 7.27e-06 & 7.06 / 62.01 / 4.61e-05 & 5.36 / 63.97 / 4.93e-06 & 7.82 \\
& RLCM & 21.02 / 80.61 / 1.99e-04 & 15.14 / 66.35 / 1.63e-04 & 14.62 / 65.66 / 1.80e-05 & 16.93 \\
& IPI & 25.67 / 85.55 / 1.15e-05 & 17.76 / 74.49 / 4.12e-05 & 27.92 / 81.37 / 1.62e-05 & 23.78 \\
\midrule
\multirow{4}{*}{Point}
& LESPS & 61.13 / 93.16 / 1.19e-05 & 58.37 / 93.76 / 2.80e-05 & 49.05 / 87.54 / 1.51e-05 & 56.18 \\
& LELCM & 58.71 / 92.26 / 3.85e-05 & 58.30 / 89.46 / 2.24e-05 & 55.23 / 87.15 / 2.27e-05 & 57.41 \\
& PAL & 66.57 / 90.49 / 2.74e-05 & 73.29 / 98.10 / 2.21e-05 & 58.40 / 91.25 / 2.45e-05 & 66.09 \\
& HMG & 72.48 / 95.06 / 1.02e-05 & 73.21 / 95.56 / 2.99e-05 & 61.21 / 94.95 / 2.07e-05 & 68.97 \\
\midrule
\multirow{7}{*}{Box}
& GrabCut & 57.78 / 96.26 / 7.31e-04 & 42.07 / 68.22 / 3.94e-04 & 50.71 / 92.04 / 2.53e-04 & 50.19 \\
& Projection Loss & 36.04 / 95.10 / 3.67e-05 & 31.41 / 98.31 / 1.88e-05 & 31.68 / 92.52 / 9.85e-05 & 33.04 \\
& Box Fill & 30.13 / 79.09 / 7.22e-05 & 25.61 / 97.99 / 1.85e-05 & 28.04 / 90.82 / 5.02e-05 & 27.93 \\
& B-SNR & 63.12 / 99.02 / 2.86e-06 & 56.12 / 93.86 / 7.68e-05 & 58.39 / 89.80 / 4.11e-05 & 59.21 \\
& BoxInst (R50) & 71.87 / 93.92 / 2.34e-05 & \textbf{76.37} / 97.14 / 3.54e-06 & 58.99 / 83.85 / 1.09e-05 & \textbf{69.08} \\
& Box2Mask (R50) & 67.51 / 93.92 / 3.79e-05 & 63.96 / 98.20 / 2.17e-05 & 55.37 / 91.07 / 7.06e-06 & 62.28 \\
& \textbf{HALO} & \textbf{73.53} / 98.04 / 7.95e-07 & 67.49 / 95.13 / 2.48e-05 & \textbf{64.33} / 89.46 / 7.06e-06 & 68.45 \\
\midrule
Full & DNANet (GT mask) & 75.52 / 96.20 / 9.63e-06 & 95.07 / 98.62 / 2.76e-06 & 70.94 / 93.54 / 9.57e-06 & 80.51 \\
\bottomrule
\end{tabular}}
\parbox{\textwidth}{\footnotesize Note: DNANet is used for LESPS, LELCM, PAL, HMG, Projection Loss, Box Fill, B-SNR, HALO, and the fully supervised upper bound unless otherwise specified. R50 denotes a ResNet-50 backbone; BoxInst and Box2Mask are included as external box-supervised segmentation frameworks, so their comparison with HALO is system-level rather than strictly same-backbone. Bold mIoU values indicate the best result within the Weak (box) group for each column. $F_a$ is a pixel-level false-alarm rate, not a percentage.}
\end{table}

Within the box-supervised group, HALO achieves the highest mIoU on
NUAA-SIRST and IRSTD-1k, outperforming the DNANet-based baselines
GrabCut, Box Fill, and B-SNR. It also beats Box2Mask on all three
datasets and BoxInst on NUAA-SIRST and IRSTD-1k; only on the low-SCR
NUDT-SIRST does BoxInst score higher (76.37\% vs.~67.49\%). HALO reaches
these results with a lightweight DNANet (ResNet-18) detector, whereas
BoxInst and Box2Mask are end-to-end instance-segmentation frameworks on a
heavier ResNet-50, so the comparison is at the system level and HALO
remains on par with or ahead of them on the two real datasets. Under
tight boxes, then, HALO is competitive but not uniformly dominant; its
more important advantage, stability under noisy boxes, is analyzed below.
Relative to fully supervised DNANet, HALO is only about 2 mIoU points
lower on NUAA-SIRST, showing that in high-SCR scenes radiometric-anchor
soft labels can approach full supervision, while a clear gap remains on
NUDT-SIRST and IRSTD-1k.

HALO also keeps stable object-level detection. Its \(P_d\) is
\(98.04\%\), \(95.13\%\), and \(89.46\%\) across the three datasets, with
\(F_a\) of \(7.95\times10^{-7}\), \(2.48\times10^{-5}\), and
\(7.06\times10^{-6}\), all low. Against the point-supervised band, HALO is
comparable in \(P_d\) and \(F_a\) on NUAA-SIRST, lower in \(F_a\) on
IRSTD-1k, and slightly weaker on NUDT-SIRST, so its gains do not come from
over-predicting foreground.

Overall, HALO's performance tracks radiometric separability. Because it
localizes from a local radiometric anchor rather than box geometry, it
stays stable under loose or center-shifted boxes; when target-background
contrast is too weak, the anchor becomes unreliable, which explains the
drop on the lower-contrast NUDT-SIRST, where BoxInst instead benefits
from clean synthetic tight boxes. This operating-regime view motivates
the mechanism ablation below and the SCR and perturbation analyses that
follow.

\paragraph{Mechanism Validation.}

We validate the two stages through controlled ablations that fix the
detector, training protocol, and data split and vary only the
box-to-label strategy, forming a progressive chain: Box Fill labels all
in-box pixels as foreground (no purification); B-SNR enables only Stage I,
shrinking supervision to radiometrically salient regions; full HALO adds
the Stage II PAG soft label.

Fig.~\ref{fig:qualitative}(b) shows the supervision improving along this
chain. Box Fill leaks substantial in-box background. B-SNR narrows to
local high-response regions, but its support is too small to cover the
target's principal-energy region. Full HALO uses PAG to rebuild smooth
support around the anchor, giving supervision closest to the true target.

The same trend appears quantitatively in Table~\ref{tab:main_results}.
Over Box Fill, Stage I (B-SNR) raises mIoU by \(32.99\), \(30.51\), and
\(30.35\) points across the three datasets, confirming that radiometric
saliency suppresses in-box background. Adding Stage II, full HALO further
improves over B-SNR by \(10.41\), \(11.37\), and \(5.94\) points,
respectively, surpassing both variants on all three datasets. PAG thus
turns the localized response into spatially smooth supervision, which is
critical to final performance and keeps the two-stage scheme robust
under different SCR regimes.

Having separated the contributions of the two stages, we next examine
why their reliability varies across datasets by analyzing the intrinsic
SCR regime and its physical interpretation.

\subsection{Operating Regime and Physical
Boundary}\label{operating-regime-and-physical-boundary}

\subsubsection{Intrinsic Contrast Differences and Operating-Regime
Analysis}\label{intrinsic-contrast-differences-and-operating-regime-analysis}

To quantify HALO's operating boundary, we compute SCR statistics over
analysis-time context windows with \(R=1.5\), \(2.0\), \(2.5\), \(3.0\),
and \(3.5\), where SCR is the local target-background contrast and \(R\)
is the context-box expansion factor. For each target component, SCR is
computed using the target mean and the background mean and standard
deviation in the surrounding context after excluding all GT target
pixels; the image-level SCR is the minimum component-level SCR. This
pure-background protocol avoids inflating the background statistics with
target pixels. Table~\ref{tab:scr_window} shows that small windows
(\(R=1.5/2.0/2.5\)) separate the datasets poorly by SCR, while the
differences stabilize at \(R=3.0\) and \(R=3.5\); we therefore use
\(R=3.0\) for the SCR analysis. This window serves only the boundary
analysis and is unrelated to the default \(r=1.5\) used for pseudo-label
generation.

\begin{table}[t]
\caption{Mean pure-background SCR under different analysis windows.}
\label{tab:scr_window}
\centering
\renewcommand{\arraystretch}{1.2}
\begin{tabular}{@{}lccccc@{}}
\toprule
Dataset & $R=1.5$ & $R=2.0$ & $R=2.5$ & $R=3.0$ & $R=3.5$ \\
\midrule
NUAA-SIRST & 6.713 & 6.889 & 7.265 & 8.141 & 8.617 \\
NUDT-SIRST & 5.011 & 5.103 & 5.240 & 5.506 & 5.600 \\
IRSTD-1k & 5.991 & 6.432 & 7.073 & 7.931 & 8.379 \\
\bottomrule
\end{tabular}
\end{table}

\begin{figure}
\centering
\includegraphics[width=\linewidth]{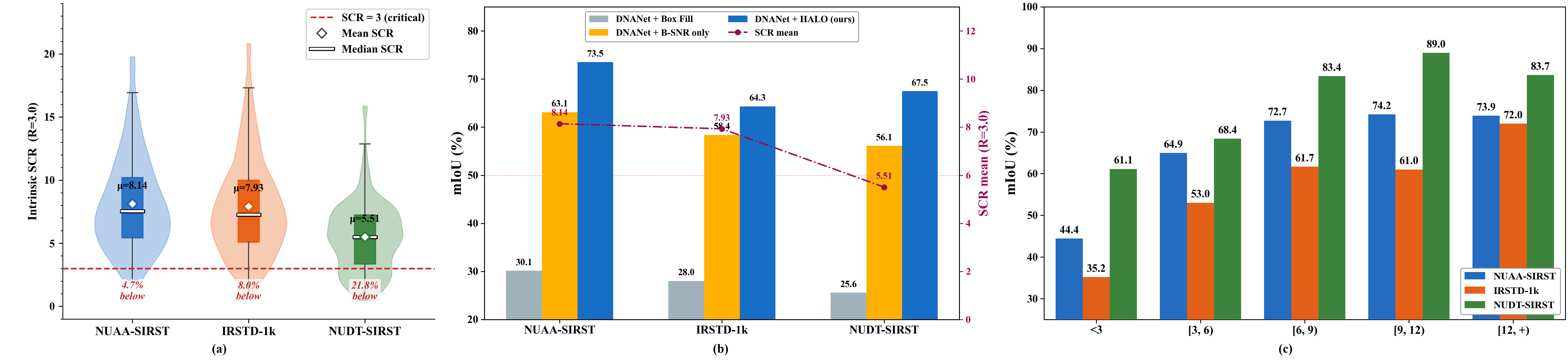}
\caption{HALO operating regime and physical boundary. (a) Intrinsic pure-background SCR distributions of the three datasets (violin plots with embedded box plots; diamonds and bars denote the mean and median SCR), where the dashed line marks the $\mathrm{SCR}=3$ separability threshold and the annotation gives the proportion of samples below it. (b) Segmentation mIoU of DNANet trained with Box Fill, B-SNR only, and full HALO supervision, overlaid with each dataset's mean SCR (right axis). (c) Average per-image mIoU of HALO (DNANet) within different intrinsic-SCR bins.}
\label{fig:regime}
\end{figure}

Fig.~\ref{fig:regime}(a) visualizes the intrinsic SCR distributions,
using \(\mathrm{SCR}=3\), a common local-contrast reference in IRSTD, as
the separability threshold \cite{rohling1983cfar}.
Table~\ref{tab:scr_stats} quantifies the differences: NUDT-SIRST has the
lowest mean SCR and the highest proportion of low-SCR (\(<3\)) samples,
21.8\%, versus 4.7\% for NUAA-SIRST and 8.0\% for IRSTD-1k, placing it
closest to the low-SCR regime where HALO is more likely to degrade.

\begin{table}[t]
\caption{Pure-background intrinsic SCR statistics under the $R=3.0$ analysis window.}
\label{tab:scr_stats}
\centering
\renewcommand{\arraystretch}{1.2}
\begin{tabular}{@{}lccccc@{}}
\toprule
Dataset & Mean & Median & Std. & SCR $<1$ (\%) & SCR $<3$ (\%) \\
\midrule
NUAA-SIRST & 8.141 & 7.534 & 3.804 & 0.2 & 4.7 \\
IRSTD-1k & 7.931 & 7.255 & 4.157 & 1.1 & 8.0 \\
NUDT-SIRST & 5.506 & 5.488 & 2.897 & 2.7 & 21.8 \\
\bottomrule
\end{tabular}
\end{table}

To link these distributions to performance, we bin test images by target
SCR and compute HALO's average per-image mIoU with DNANet, as shown in
Fig.~\ref{fig:regime}(b,c). Across all three datasets mIoU rises with
SCR and inflects around \(\mathrm{SCR}\approx 3\). In the \([3,6)\) and
\([6,9)\) bins, NUAA-SIRST reaches \(64.9\%\) and \(72.7\%\), NUDT-SIRST
\(68.4\%\) and \(83.4\%\), and IRSTD-1k \(53.0\%\) and \(61.7\%\); in the
\(\mathrm{SCR}<3\) bin, mIoU drops to \(44.4\%\) (NUAA-SIRST), \(35.2\%\)
(IRSTD-1k), and \(61.1\%\) (NUDT-SIRST). NUDT-SIRST has the highest per-bin
mIoU everywhere, yet its overall mIoU (\(67.49\%\)) trails NUAA-SIRST
(\(73.53\%\)): it simply holds many more low-SCR samples (\(21.8\%\) below
\(\mathrm{SCR}=3\)), so its dataset average is pulled down by hard cases,
not by weaker behavior at a given SCR. HALO's low-SCR sensitivity is thus
tied to the SCR regime, not to a particular dataset. Below about
\(\mathrm{SCR}=3\) the target is poorly separable from local background
and the B-SNR anchor is prone to degradation, so \(\mathrm{SCR}\approx 3\)
marks an approximate critical threshold for reliable HALO operation.

\subsubsection{Physical Mechanism
Interpretation}\label{physical-mechanism-interpretation}

To explain the relationship among context-statistics contamination,
intrinsic SCR, and B-SNR anchor reliability, we derive a simplified
expression for the variance of the context window \(\Omega_{ctx}\).
Under box supervision, \(\Omega_{ctx}\) is a mixture of two pixel
groups: target pixels with proportion \(\alpha\), mean \(\mu_t\), and
variance \(\sigma_t^2\), and background pixels with proportion
\(1-\alpha\), mean \(\mu_b\), and variance \(\sigma_b^2\). Let the mean
difference be \(\Delta\mu=\mu_t-\mu_b\). By the law of total variance,
the context variance is

\[
\hat{\sigma}_{\mathrm{ctx}}^2 = \big[\alpha\,\sigma_t^2 + (1-\alpha)\,\sigma_b^2\big] + \alpha(1-\alpha)\,\Delta\mu^2 ,
\]

where the first bracket is the within-group variance and the second is
the between-group variance from their mean difference. If the two groups
have similar internal dispersion, both close to the ideal pure-background
variance \(\sigma_0^2\), the within-group term reduces to \(\sigma_0^2\),
giving

\[
\hat{\sigma}_{\mathrm{ctx}}^2 \approx \sigma_0^2 + \alpha(1-\alpha)\Delta\mu^2 ,
\]

which holds when \(\Omega_{ctx}\) is a mixture of two roughly homogeneous
groups and \(\alpha\) is the main source of variance inflation. The
contamination term \(\alpha(1-\alpha)\Delta\mu^2\) is non-negative, peaks
at \(\alpha=0.5\), and vanishes as \(\alpha\to 0\) or \(1\); target pixels
leaking into the window therefore inflate the estimated background
variance, most strongly under moderate contamination.

This links to intrinsic SCR: while \(\mathrm{SCR}\approx
\Delta\mu/\sigma_0\), the anchor's reliability depends on the effective
saliency under the actual normalized background,
\(\Delta\mu/\hat{\sigma}_{\mathrm{ctx}}\). At low SCR with strong
contamination, \(\arg\max\) tends to degenerate into noise localization;
at high SCR with weak contamination, the anchor is stable and PAG can
match the true diffusion range.

This matches the data: on low-SCR NUDT-SIRST, DNANet+B-SNR improves
substantially over Box Fill (Table~\ref{tab:main_results}) but still
remains below full HALO, showing that the Stage-I anchor alone is
insufficient under low SCR and strong contamination. PAG mitigates part
of this instability but is still bounded by the Stage-I anchor's
reliability, so the same contamination-aware analysis explains both
HALO's strengths and its limits.

To examine the design choice of replacing the geometric center with the
radiometric anchor \(p^\star\), Table~\ref{tab:anchor_ablation} compares the two strategies
under tight boxes.

\begin{table}[t]
\caption{Ablation comparison between the radiometric anchor \(p^\star\) and the geometric center.}
\label{tab:anchor_ablation}
\centering
\renewcommand{\arraystretch}{1.2}
\begin{tabular}{@{}llcc@{}}
\toprule
Dataset & Anchor choice & mIoU (\%) & \(\Delta\) vs. center \\
\midrule
\multirow{2}{*}{NUAA-SIRST}
  & Radiometric peak \(p^\star\) & 73.53 & +3.99 \\
  & Box center & 69.54 & -- \\
\multirow{2}{*}{NUDT-SIRST}
  & Radiometric peak \(p^\star\) & 67.49 & -0.99 \\
  & Box center & 68.48 & -- \\
\multirow{2}{*}{IRSTD-1k}
  & Radiometric peak \(p^\star\) & 64.33 & +3.99 \\
  & Box center & 60.34 & -- \\
\bottomrule
\end{tabular}
\end{table}

The radiometric anchor's benefit depends on the operating regime. Under
tight boxes the target is already well enclosed and the geometric center
usually sits on it, so the gain is limited: about \(4\) mIoU points on
NUAA-SIRST and IRSTD-1k, and essentially a tie on low-SCR NUDT-SIRST
(\(67.49\%\) vs.~\(68.48\%\)). The anchor's value lies less in tight-box
accuracy than in not depending on precise box geometry: a geometric
center drifts when boxes loosen or shift, whereas the radiometric anchor
tracks the strongest in-box response. Its advantage should therefore be
clearest under loose or shifted boxes, consistent with the box-expansion
results below.

\begin{figure}
\centering
\includegraphics[width=0.9\linewidth]{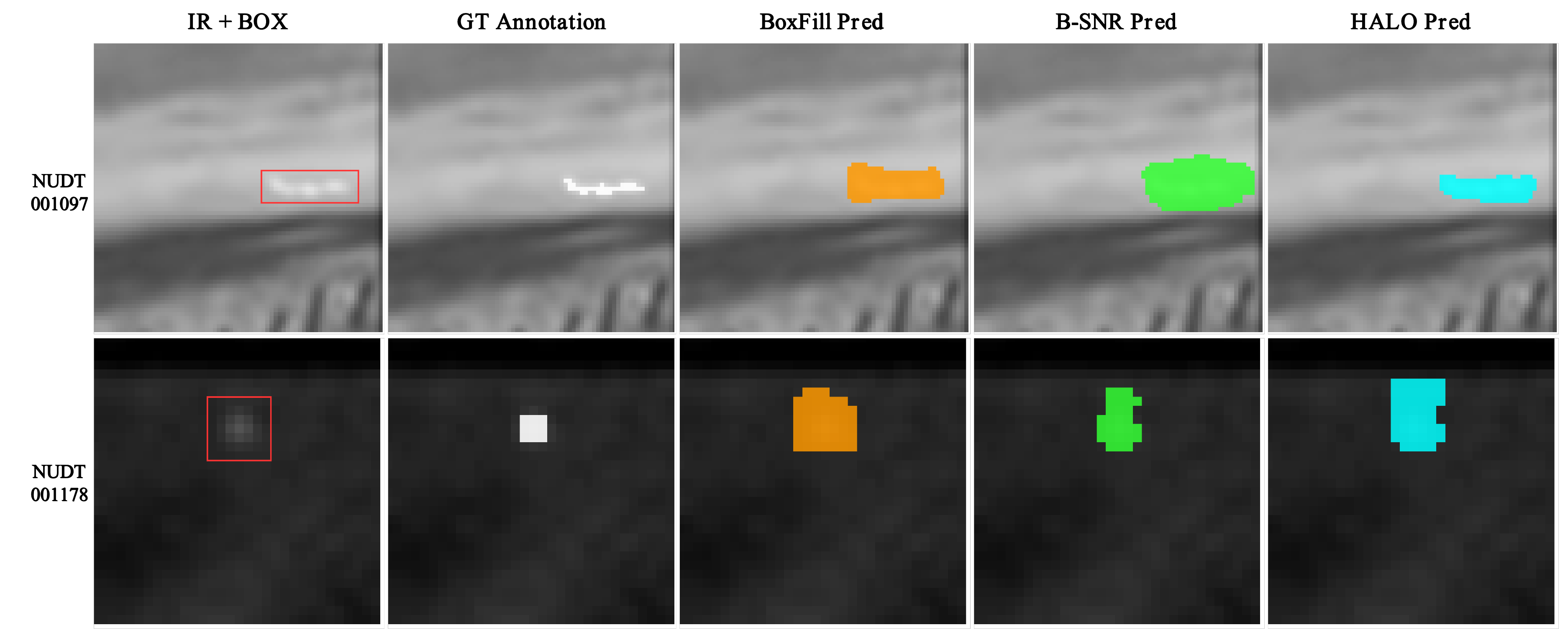}
\caption{Typical HALO failure cases under low-SCR scenarios. Each row is one example (NUDT-001097 and NUDT-001178); from left to right, the columns show the infrared image with the box annotation, the ground-truth mask, and the predictions of DNANet trained with Box Fill, B-SNR, and HALO supervision. Under weak local contrast, the predicted target region over-expands beyond the true target.}
\label{fig:failure}
\end{figure}

Fig.~\ref{fig:failure} shows two typical low-SCR failures. As local SCR
drops, the B-SNR peak becomes less separable, so the estimated
\(\Omega_{1/2}\) and \(\sigma_g\) deviate from the true target extent and
both the soft label and the prediction over-expand.

\subsection{Box-Perturbation Robustness}\label{box-perturbation-robustness}

\begin{figure}
\centering
\includegraphics[width=0.8\linewidth]{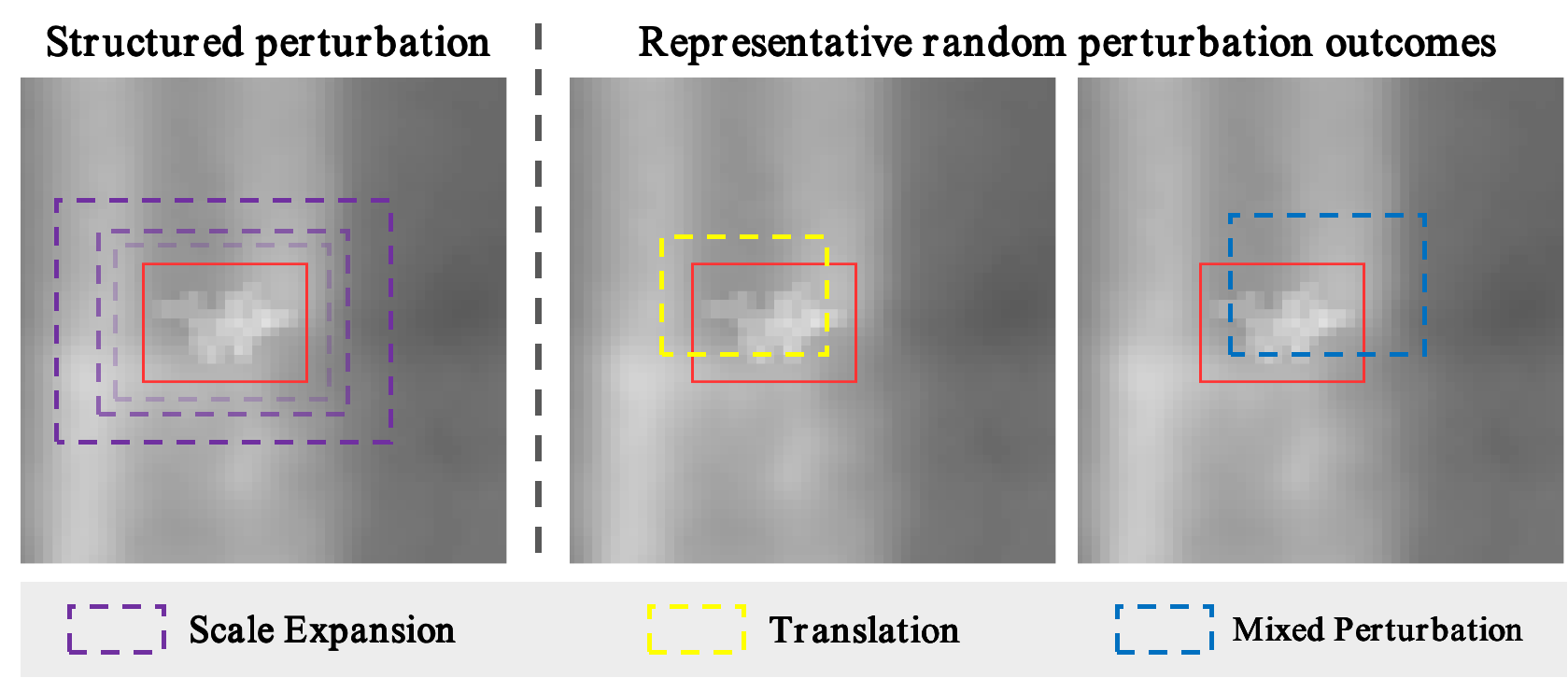}
\caption{Illustration of the two box perturbation types: structured scale expansion and random four-sided perturbation.}
\label{fig:perturbation_diagram}
\end{figure}

Real-world boxes often have loose boundaries or localization shifts. We
test HALO's robustness at two levels. At the pseudo-label level we measure
IoU retention against ground-truth masks under perturbed boxes; at the
end-to-end level we train downstream models on the perturbed-box labels
and compare mIoU with generic box-supervised methods. Two perturbation
types, shown in Fig.~\ref{fig:perturbation_diagram}, are used: structured scale
expansion, simulating oversized detector candidates, and random
four-sided perturbation, which samples offsets independently on the four
sides.

\begin{figure}
\centering
\includegraphics[width=0.95\textwidth]{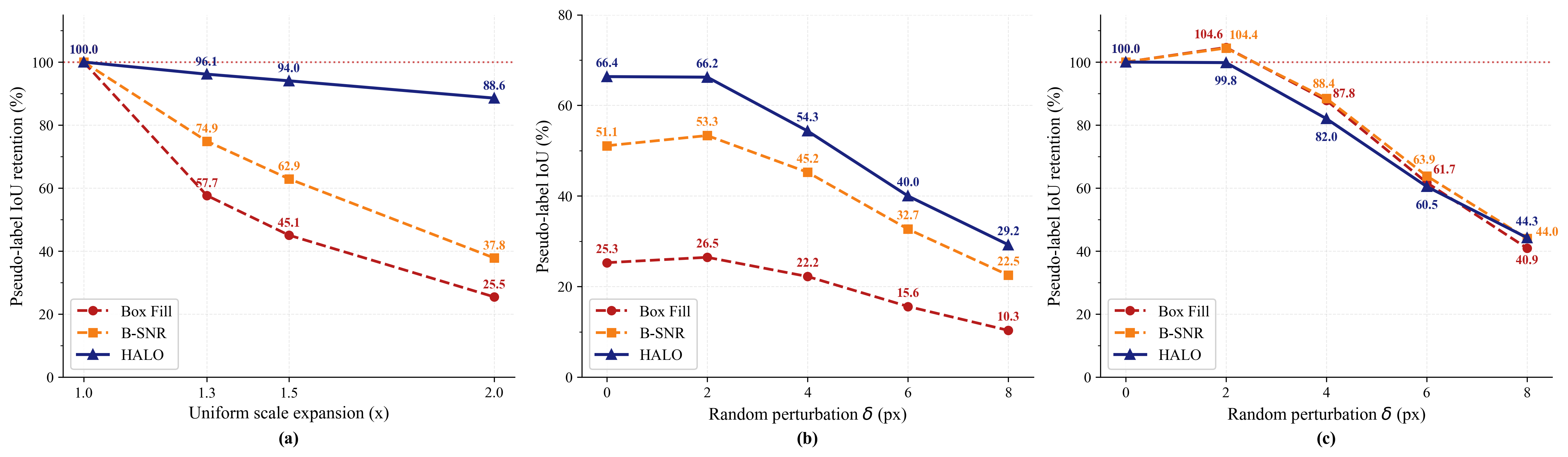}
\caption{Robustness of HALO to structured and random box perturbations at the pseudo-label level, for Box Fill, B-SNR, and HALO. (a) Pseudo-label IoU retention under uniform scale expansion. (b) Absolute pseudo-label IoU under random four-sided perturbation of magnitude $\delta$. (c) Pseudo-label IoU retention under the same random perturbation.}
\label{fig:perturbation_robustness}
\end{figure}

At the pseudo-label level, Fig.~\ref{fig:perturbation_robustness} shows that
expanding boxes from \(1.0\times\) to \(2.0\times\) leaves Box Fill,
B-SNR, and HALO with IoU retention of \(25.5\%\), \(37.8\%\), and
\(88.6\%\), so HALO degrades only slightly. Under random four-sided
perturbation it barely degrades at \(\delta=2\) px, and even at
\(\delta=8\) px stays about \(18\)-\(20\) points above Box Fill. Once the
center shift reaches about \(4\) px, near a typical target radius, HALO
and all baselines drop sharply, indicating that the anchor needs the box
localization error to stay below the target radius and is not guaranteed
under arbitrary shifts.

To see whether this transfers to the final task, we compare HALO with
BoxInst \cite{tian2021boxinst} and Box2Mask \cite{li2022box2mask} end to
end under pure scale expansion (Table~\ref{tab:scale_expansion}); each
method follows its standard implementation and checkpoint-selection
protocol.

\begin{table}[t]
\caption{End-to-end mIoU (\%) under pure scale expansion. Each dataset column reports \(1.0 / 2.0 / 2.5\times\) boxes.}
\label{tab:scale_expansion}
\centering
\setlength{\tabcolsep}{6pt}
\renewcommand{\arraystretch}{1.2}
\begin{tabular}{@{}lccc@{}}
\toprule
Method & NUAA-SIRST & NUDT-SIRST & IRSTD-1k \\
\midrule
Box2Mask & 67.51 / 39.84 / 36.89 & 63.96 / 16.32 / 13.83 & 55.37 / 27.66 / 18.66 \\
BoxInst & 71.87 / 62.10 / 53.54 & 76.37 / 23.24 / 9.62 & 58.99 / 51.80 / 48.64 \\
HALO & 73.53 / 61.08 / 60.99 & 67.49 / 49.01 / 36.89 & 64.33 / 56.13 / 54.08 \\
\bottomrule
\end{tabular}
\parbox{\textwidth}{\footnotesize Note: \(1.0\times\) denotes tight boxes. The best method for each dataset and expansion level varies with SCR: BoxInst is stronger under tight boxes, whereas HALO is more stable under expansion and low SCR.}
\end{table}

The end-to-end results echo the pseudo-label findings. Box2Mask collapses
as box expansion adds background, especially at \(2.0\times\) and
\(2.5\times\), while HALO declines far less. Against the stronger BoxInst
the picture depends on SCR: the two are close on high-SCR NUAA-SIRST
(\(61.08\%\) vs.~\(62.10\%\) at \(2.0\times\)), but on low-SCR NUDT-SIRST box
loosening badly hurts BoxInst's projection consistency, which falls to
\(23.24\%\)/\(9.62\%\) at \(2.0\times\)/\(2.5\times\) while HALO holds
\(49.01\%\)/\(36.89\%\). Additional position-jitter tests show the same trend,
with HALO retaining a larger margin on NUDT-SIRST, but the pure
scale-expansion setting in Table~\ref{tab:scale_expansion} is used as
the primary end-to-end comparison. Because the radiometric anchor is
driven by the strongest in-box response, it is largely independent of
the box boundary and tolerates loosening that breaks the
projection-consistency and level-set assumptions of BoxInst and Box2Mask.

HALO's robustness advantage is clearest under low-SCR noisy boxes: it
matches strong generic box-supervised methods at high SCR and degrades
more slowly at low SCR with imprecise boxes. Because HALO requires only
one offline label-generation step and can be used with a lightweight
ResNet-18 detector, it is practical where annotation errors are
unavoidable. We compare only
box-supervised methods here; HALO's gap to point-supervised methods
appears mainly in clean low-SCR settings and does not imply that box
supervision is generally superior to point supervision.

\subsection{Additional Validation}\label{additional-validation}

After the main comparison, operating-regime analysis, and noisy-box
robustness tests, we further check whether HALO's behavior depends on
particular hyperparameter choices or on a specific detector backbone.

\paragraph{Default Settings and Parameter Sensitivity.}

\begin{figure}
\centering
\includegraphics[width=0.8\linewidth]{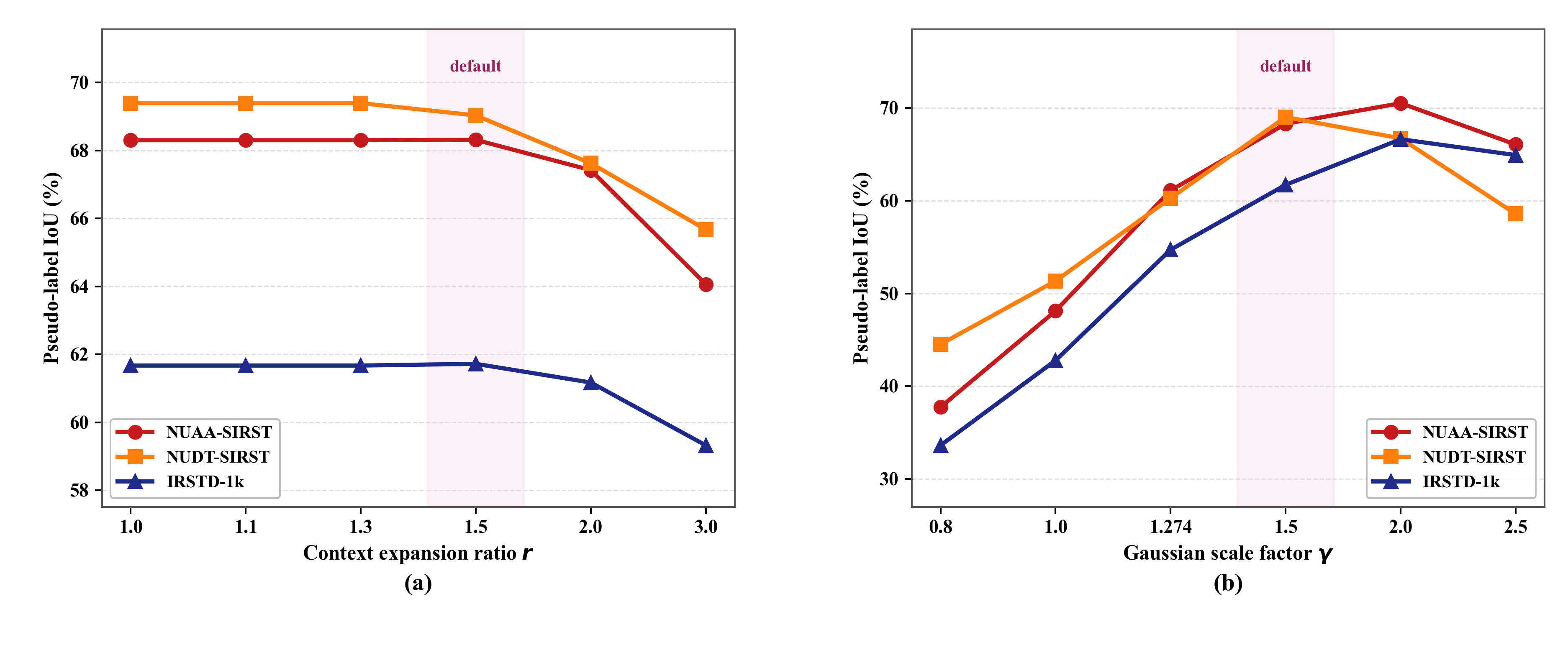}
\caption{Parameter sweep results for $r$ and $\gamma$, evaluated by pseudo-label IoU. (a) Pseudo-label IoU versus the context expansion ratio $r$. (b) Pseudo-label IoU versus the Gaussian scale factor $\gamma$. The shaded band marks the default setting $r=\gamma=1.5$.}
\label{fig:parameter_sweep}
\end{figure}

HALO has two core parameters: the context expansion ratio \(r\), which
sets how far the background window extends around each box in Stage I, and
the Gaussian scale factor \(\gamma\), which scales the estimated spread
\(\sigma_g\) of the PAG soft label in Stage II. Both are set to 1.5 by
default. Fig.~\ref{fig:parameter_sweep} reports the pseudo-label-quality
sweeps and Table~\ref{tab:default_params} the end-to-end validation. For
\(r\), pseudo-label intersection-over-union (IoU) is stable over
\(r\in[1.0,2.0]\) and only drops at \(r=3.0\), where the enlarged window
dilutes the local background statistics; end-to-end, \(r=1.5\) beats
\(r=3.0\) on all three datasets. For \(\gamma\), \(\gamma=2.0\) gives
pseudo-label IoU close to \(\gamma=1.5\) on some datasets, but a larger
spread over-smooths the label and lowers end-to-end mIoU, which is
consistently best at \(\gamma=1.5\). HALO is thus insensitive to both
parameters over a wide range, so they can be fixed as constants from
imaging and numerical considerations instead of being tuned per dataset.

\begin{table}[t]
\caption{Validation of the default parameter choices using end-to-end mIoU (\%).}
\label{tab:default_params}
\centering
\setlength{\tabcolsep}{6pt}
\renewcommand{\arraystretch}{1.2}
\begin{tabular}{@{}lcccc@{}}
\toprule
Parameter & Values & NUAA-SIRST & NUDT-SIRST & IRSTD-1k \\
\midrule
\(r\) & 1.5 / 3.0 & 73.53 / 70.11 & 67.49 / 61.50 & 64.33 / 58.03 \\
\(\gamma\) & 1.5 / 2.0 & 73.53 / 69.21 & 67.49 / 64.27 & 64.33 / 62.57 \\
\bottomrule
\end{tabular}
\end{table}

\FloatBarrier

\paragraph{Cross-Backbone Compatibility.}

To check that HALO does not depend on a specific backbone, we evaluate
four detectors beyond DNANet: SCTransNet \cite{yuan2024sctransnet},
MSHNet \cite{liu2024mshnet}, ALCNet, and ACM, each under GT-mask, B-SNR,
and HALO supervision (mIoU / \(P_d\) / \(F_a\)). Table~\ref{tab:backbone}
reports them; DNANet appears in Table~\ref{tab:main_results}.

\begin{table}[t]
\caption{Cross-backbone compatibility of HALO. Each entry reports mean intersection-over-union (mIoU, \%) / probability of detection (\(P_d\), \%) / false-alarm rate (\(F_a\)).}
\label{tab:backbone}
\centering
\footnotesize
\setlength{\tabcolsep}{4pt}
\renewcommand{\arraystretch}{1.2}
\resizebox{\textwidth}{!}{%
\begin{tabular}{@{}llccc@{}}
\toprule
Dataset & Backbone & GT mask & B-SNR & HALO \\
\midrule
\multirow{4}{*}{NUAA-SIRST}
  & SCTransNet & 74.94 / 97.72 / 2.35e-05 & 61.26 / 94.68 / 1.04e-04 & \textbf{71.47 / 95.06 / 1.57e-05} \\
  & MSHNet & 71.67 / 96.20 / 3.69e-05 & 59.93 / 93.54 / 2.47e-05 & \textbf{68.55 / 95.06 / 1.75e-05} \\
  & ALCNet & 69.72 / 97.06 / 1.56e-05 & 59.31 / 97.06 / 1.00e-05 & \textbf{66.93 / 96.08 / 1.43e-06} \\
  & ACM & 68.95 / 97.06 / 2.54e-06 & 53.80 / 96.08 / 3.18e-06 & \textbf{66.28 / 98.04 / 4.77e-07} \\
\multirow{4}{*}{NUDT-SIRST}
  & SCTransNet & 84.31 / 98.31 / 5.49e-06 & 56.78 / 94.07 / 4.69e-05 & \textbf{67.79 / 95.03 / 2.37e-05} \\
  & MSHNet & 80.29 / 97.14 / 1.19e-05 & 52.72 / 93.97 / 3.27e-05 & \textbf{64.40 / 95.87 / 3.53e-05} \\
  & ALCNet & 66.16 / 94.60 / 1.96e-05 & 51.95 / 93.33 / 4.82e-05 & \textbf{55.81 / 94.50 / 3.29e-05} \\
  & ACM & 64.98 / 95.66 / 2.52e-05 & 51.27 / 92.70 / 5.47e-05 & \textbf{54.25 / 93.12 / 3.75e-05} \\
\multirow{4}{*}{IRSTD-1k}
  & SCTransNet & 67.12 / 87.76 / 7.90e-06 & 59.56 / 91.16 / 1.26e-05 & \textbf{65.03 / 91.50 / 2.48e-05} \\
  & MSHNet & 66.21 / 91.84 / 1.52e-05 & 56.32 / 89.12 / 2.73e-05 & \textbf{63.28 / 91.50 / 9.79e-06} \\
  & ALCNet & 58.44 / 84.35 / 2.73e-05 & 54.07 / 82.31 / 1.04e-05 & \textbf{55.90 / 87.76 / 3.67e-05} \\
  & ACM & 57.79 / 85.03 / 2.06e-05 & 52.55 / 85.03 / 3.49e-05 & \textbf{54.45 / 83.33 / 3.32e-05} \\
\bottomrule
\end{tabular}}
\parbox{\textwidth}{\footnotesize Note: GT mask denotes the fully supervised upper bound and is not included in the bold marking. Bold entries indicate the better mIoU within the weak-supervision comparison between B-SNR and HALO.}
\end{table}

HALO pairs with all four backbones. On the recent SCTransNet and MSHNet
it reaches mIoU comparable to DNANet, so the supervision-generation
strategy is not tied to a particular architecture. It consistently beats
the corresponding B-SNR variant, and approaches GT-mask supervision on
NUAA-SIRST and IRSTD-1k, with a larger gap on NUDT-SIRST that again
follows the low-SCR boundary.

\subsection{Efficiency and
Practicality}\label{efficiency-and-practicality}

The preceding results show that HALO is decoupled from the choice of
detector backbone. We therefore also evaluate its computational footprint
from the perspective of efficiency and engineering practicality. HALO
does not introduce a complex additional pipeline: its pseudo-labels are
generated once offline before training and are independent of the
current state of the detector backbone. Across 1,687 training
images from the three datasets, full soft-label generation takes only
0.43\,s on a single CPU process, or about
0.25\,ms/image. This process requires no GPU inference or
additional network forward pass. These timings characterize the offline
label-generation overhead rather than a hardware-normalized runtime
benchmark. By contrast, LESPS, PAL, and HMG
generate or refine their pseudo-labels online during training, so that
label updates are coupled to the network state and repeated throughout
optimization. HALO instead decouples label generation from training
entirely, adding essentially no online overhead.

Under a \(256\times256\) single-channel input and a batch size of 1,
Table~\ref{tab:efficiency} summarizes the parameter count, computational cost in
multiply-accumulate operations (MACs), and inference speed in frames per
second (FPS) of five backbones. ACM and ALCNet are the lightest, with
about \(0.4\)-\(0.5\) million parameters and fewer than \(0.5\) G MACs,
and also provide the fastest inference. MSHNet is intermediate. DNANet
has the highest computational cost, whereas SCTransNet, as a transformer
backbone, has the largest parameter count. Because HALO changes only
offline supervision generation and does not alter the downstream network
structure, its inference-time parameters, MACs, and FPS are fully
determined by the selected detection backbone. It therefore introduces
no additional online computation. The method can thus be paired with
lightweight backbones such as ACM/ALCNet or higher-capacity backbones
such as DNANet/SCTransNet according to the available compute budget.

\begin{table}[t]
\caption{Backbone efficiency comparison under \(256\times256\) input.}
\label{tab:efficiency}
\centering
\setlength{\tabcolsep}{10pt}
\renewcommand{\arraystretch}{1.2}
\begin{tabular}{@{}lccc@{}}
\toprule
Backbone & \#Params (M) & MACs (G) & FPS \\
\midrule
ACM & 0.40 & 0.40 & 386.0 \\
ALCNet & 0.52 & 0.38 & 421.7 \\
MSHNet & 4.07 & 6.10 & 127.4 \\
DNANet & 4.70 & 14.26 & 57.3 \\
SCTransNet & 11.33 & 10.12 & 58.6 \\
\bottomrule
\end{tabular}
\end{table}

\section{Conclusion}\label{sec:conclusion}

In this paper, we proposed HALO to transform box annotations into soft
supervision for pixel-level training in box-supervised IRSTD. HALO
localizes a physically grounded radiometric anchor inside each box and
synthesizes a Physically Anchored Gaussian soft label around it,
converting highly noisy boxes into continuous supervision offline,
decoupled from the detector and without online label updates.

Experiments on NUAA-SIRST, NUDT-SIRST, and IRSTD-1k support three
findings. First, under standard tight boxes HALO is competitive with
representative box-supervised methods and, on high-SCR NUAA-SIRST, comes
within about two mIoU points of full mask supervision. Second, its main
advantage is robustness: under \(2.0\times\)/\(2.5\times\) box expansion
and four-sided jitter, HALO degrades far more slowly than BoxInst and
Box2Mask, whose projection and level-set priors collapse once the
tight-box assumption breaks. Third, the contamination-aware
operating-regime analysis ties this behavior to intrinsic signal-to-clutter
ratio, identifying \(\mathrm{SCR}\approx 3\) as an approximate reliability
threshold and showing that the low-SCR sensitivity of HALO is a property
of the SCR regime rather than of any single dataset. HALO also transfers
across five detector backbones at negligible inference cost.

In practice, when coarse boxes are already available, HALO offers a
low-cost path to trainable pixel-level supervision and reduces the
reliance of infrared small-target segmentation on dense mask annotation.
Its main limitation is inherited from the Stage-I anchor: under ultra-low
SCR or box shifts larger than the target radius, the radiometric peak
becomes unreliable and the soft label can over-expand. Future work may
explore adaptive anchor correction for such regimes, cross-dataset and
cross-sensor generalization, and extensions to multimodal box-supervised
infrared detection.

\FloatBarrier

\section*{Acknowledgements}
This work was supported by the Tianjin Natural Science Foundation Project (23JCJQJC00070) and the National Natural Science Foundation of China (NSFC) (62272342, T2422015).

\section*{Declaration of competing interest}
The authors declare that they have no known competing financial interests or personal relationships that could have appeared to influence the work reported in this paper.

\section*{Data availability}
The datasets used in this study (NUAA-SIRST, IRSTD-1k, and NUDT-SIRST) are publicly available.

\section*{Declaration of generative AI and AI-assisted technologies in the writing process}
During the preparation of this work, the authors used a generative AI tool to improve the language and readability of the manuscript. The authors reviewed and edited the output as needed and take full responsibility for the content of the published article.

\section*{CRediT authorship contribution statement}
\textbf{Xizhe Zhang:} Conceptualization, Methodology, Software, Validation, Investigation, Visualization, Writing -- original draft. \textbf{Fan Shi:} Conceptualization, Formal analysis, Resources, Supervision, Funding acquisition, Writing -- review \& editing. \textbf{Mianzhao Wang:} Methodology, Investigation, Validation, Writing -- review \& editing. \textbf{Jiangpeng Zheng:} Investigation, Validation, Writing -- review \& editing. \textbf{Xu Cheng:} Supervision, Writing -- review \& editing. \textbf{Shengyong Chen:} Supervision, Resources, Funding acquisition.

\bibliographystyle{elsarticle-num}
\bibliography{IEEEabrv,IEEEexample}

\end{document}